\RequirePackage{fix-cm}
\documentclass[twocolumn]{svjour3}          
\smartqed  
\usepackage{graphicx}
\usepackage[]{color}
\usepackage[]{amsmath}
\usepackage{amssymb}
\usepackage{textcomp}
\usepackage{subfigure}
\usepackage{multirow}
\usepackage{hyphenat} 

\graphicspath{ {figs/res/}{figs/nlro/} }

\newcommand{\e}[1]{\times 10^{#1}}
\newcommand{\units}[1]{\; \text{#1}}

\definecolor{rev1}{rgb}{0,0,0}
\definecolor{rev2}{rgb}{0,0,0}
\definecolor{orange}{rgb}{1,0.5,0}

\begin{document}

\title{Online training for high-performance analogue readout layers in photonic reservoir computers}


\author{Piotr Antonik \and Marc Haelterman \and Serge Massar}


\institute{P. Antonik, S. Massar \at
           Laboratoire d'Information Quantique \\ 
           Universit\'{e} libre de Bruxelles \\
           Avenue F. D. Roosevelt 50, CP 224 \\
           Brussels, Belgium \\
           \email{pantonik@ulb.ac.be}           
           \and
           M. Haelterman \at
           Service OPERA-Photonique \\
           Universit\'{e} libre de Bruxelles, \\
           Avenue F. D. Roosevelt 50, CP 194/5 \\
           Brussels, Belgium
}

\date{}

\maketitle

\begin{abstract}
  \emph{Introduction.} Reservoir Computing is a bio-inspired computing paradigm for processing time\hyp{}dependent signals. The performance of its hardware implementation is comparable to state-of-the-art digital algorithms on a series of benchmark tasks. The major bottleneck of these implementation is the readout layer, based on slow offline post-processing. Few analogue solutions have been proposed, but all suffered from noticeable decrease in performance due to added complexity of the setup. \\
  \emph{Methods.} 
  Here we propose the use of online training to solve these issues. We study the applicability of this method using numerical simulations of an experimentally feasible reservoir computer with an analogue readout layer. We also consider a nonlinear output layer, which would be very difficult to train with traditional methods. \\
  \emph{Results.} We show numerically that online learning allows to circumvent the added complexity of the analogue layer and obtain the same level of performance as with a digital layer. \\
  \emph{Conclusion.} 
  This work paves the way to high\hyp{}performance fully-analogue reservoir computers through the use of online training of the output layers.
\keywords{Reservoir computing \and Opto-electronics \and Analogue readout \and FPGA \and Online training}
\end{abstract}

\section{Introduction}
\label{sec:intro}

Reservoir computing is a set of machine learning methods for designing and training artificial neural networks, introduced independently in \cite{jaeger2004harnessing} and \cite{maass2002real}. The idea is that one can exploit the dynamics of a recurrent nonlinear network to process time series without training the network itself, but simply adding a general linear readout layer and only training the latter. This results in a system that is significantly easier to train (the learning is reduced to solving a system of linear equations, see \cite{lukovsevivcius2009survey}), yet powerful enough to match other algorithms on a series of benchmark tasks. Reservoir computing has been successfully applied to wireless channel equalisation and chaotic time series forecasting \cite{jaeger2004harnessing}, phoneme recognition \cite{triefenbach2010phoneme}, image processing \cite{meftah2016novel}, handwriting recognition \cite{malik2014novel}, audio classification \cite{scardapane2016semi} and won an international competition on prediction of future evolution of financial time series \cite{NFC}. 

Reservoir computing allows efficient implementation of simplified recurrent neural networks in hardware, such as e.g. optical components. Optical computing has been investigated for decades as photons propagate faster than electrons, without generating heat or magnetic interference, and thus promise higher bandwidth than conventional computers \cite{arsenault2012optical}. Reservoir computing would thus allow to build high-speed and energy efficient photonic \textcolor{rev2}{computational} devices. Several important steps have been taken towards this goal with electronic \cite{appeltant2011information}, opto-electronic \cite{paquot2012optoelectronic,larger2012photonic,martinenghi2012photonic}, all-optical \cite{duport2012all,brunner2012parallel,vinckier2015high} and integrated \cite{vandoorne2014experimental} experimental implementation reported since \textcolor{rev2}{2011}.

The major drawback in these experiments is the absence of efficient readout mechanisms: the states of the neurons are collected and post-processed on a computer, severely reducing the processing speeds and thus limiting the applicability. An analog readout would resolve this issue, as suggested in \cite{woods2012optical}. This research direction has already been investigated experimentally in \cite{smerieri2012analog,duport2016fully,vinckier2016autonomous}, but all these implementations suffered from significant performance degradation due to the complex structure of the readout layer. 
\textcolor{rev2}{Indeed the approach used in these works was to characterise with high accuracy the linear output layer, whereupon it was possible to compute offline the output weights. However it is virtually impossible to characterise each hardware component of the setup with sufficient level of accuracy. Furthermore the components in the output layer may have a slight nonlinear behaviour. It follows that this approach does not work satisfactorily, as is apparent from the performance degradation reported in \cite{duport2016fully}.}

In this work we address the above issues with the online learning approach. \textcolor{rev2}{Online training has attracted much attention in the machine learning community because it allows to train the system gradually, as the input data becomes available. It can also easily cope with non-stationary input signal, whose characteristics change with time, as the online approach can keep the model updated according to variations in the input. Finally, in the case of hardware systems, online training can easily cope with drifts in the hardware, as the system will adapt to gradual changes in the hardware components \cite{bottou1998online,shalev2012online}.}

\textcolor{rev2}{In the context of reservoir computing, the online training implements a gradient descent: it gradually changes the output layer to adapt to the task. More precisely the output layer is characterised by a series of parameters (the readout weights), and in online training these weights are adjusted in small increments, so that the output of the system gets closer to the target signal.
  We have previously applied this method to a hardware reservoir computer with a digital output layer in \cite{antonik2016online}, where we illustrated how online learning could cope with non-stationary input signals, i.e. tasks that change with time.}

\textcolor{rev2}{The important point in the present context is that, compared to previously used offline methods, in online training based on gradient descent no assumption is necessary about how these weights contribute to the output signal. That is, it is not necessary to model the output layer. Furthermore, the transfer function of the readout layer could in principle be nonlinear. Here we show, using realistic numerical simulations, how these features could be highly advantageous for training hardware reservoir computers.}

\textcolor{rev2}{For concreteness, we will consider in simulations an opto-electronic reservoir computing setup based on a ring topology already extensively studied experimentally in \cite{paquot2012optoelectronic,larger2012photonic}. We add to this setup an analogue layer that is now trained online by an FPGA chip processing the simple gradient descent algorithm in real time, as in \cite{antonik2016online}.} The readout layer consists of a simple Resistor-Capacitor (RC) circuit (as in \cite{smerieri2012analog}), instead of a more complicated RLC circuit (consisting of a resistor $R$, an inductor $L$ and a capacitor $C$) that was used to increase the amplitude of the output signal in \cite{duport2016fully}.

\textcolor{rev2}{We} investigate the performance of this setup through numerical simulations on two benchmark tasks and show that previously encountered difficulties are almost entirely alleviated by the online training approach. In other words, with a relatively simple analogue readout layer, trained online, and without any modelling of the underlying processes, we obtain results similar to those produced by a digital layer, trained offline. We also explore a special case with a nonlinear readout function and show that this complication doesn't decrease much the performance of the system. \textcolor{rev2}{This work thus brings an interesting solution to an important problem in the hardware reservoir computing field.}

The paper is structured as follows. In the Methods section, we introduce the basic principles of reservoir computing, online learning and the benchmark tasks used here, and then present the experimental opto\hyp{}electronic reservoir computer, the analogue readout layer, and specify the major aspects of our numerical simulations. We then focus on the results of our investigations and conclude the paper with future perspectives.

\section{Methods}

\subsection{Reservoir Computing}

\begin{figure}
  \centering
  \includegraphics[width=0.45\textwidth]{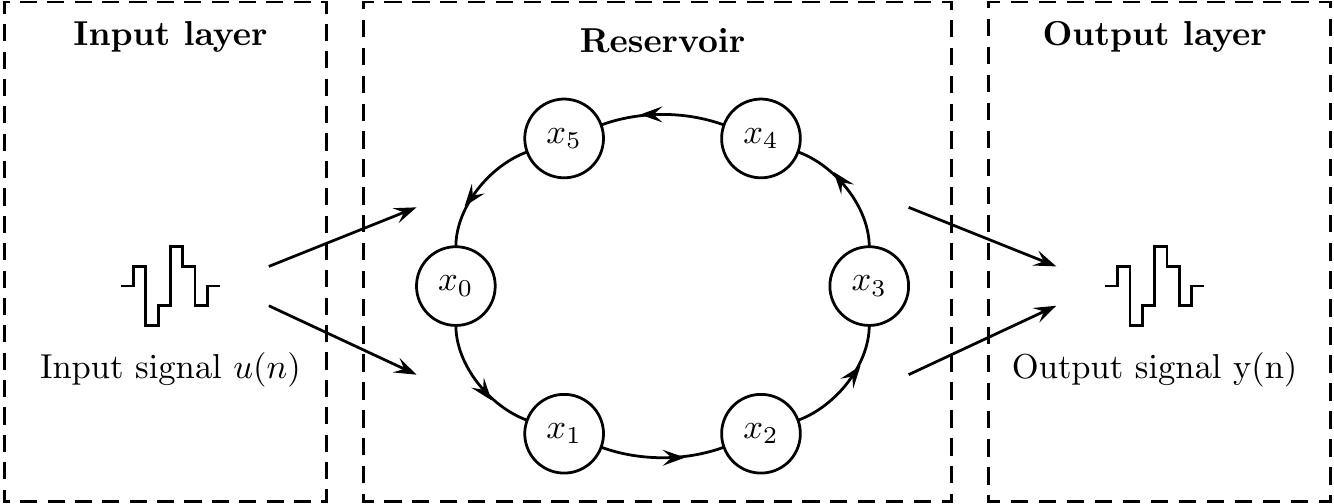}
  \caption{Schematic representation of a reservoir computer with $N=6$ nodes. In terms of artificial neural networks, its architecture is composed of a single input neuron (which receives the input signal $u(n)$), one layer of hidden neurons and a single output neuron (which produces the output signal $y(n)$). The configuration of neurons in the hidden layer can be arbitrary, but for ease of hardware implementation we use a ring-like topology.
}
  \label{fig:rc}
\end{figure}

A reservoir computer, schematised in figure \ref{fig:rc}, consists of a recurrent network of internal variables, usually called ``nodes'' or ``neurons'', from its biological origins \cite{lukovsevivcius2009survey}. These $N$ variables, denoted \textcolor{rev2}{by} $x_i(n)$, with $i=0,\ldots,N-1$, evolve in discrete time $n \in \mathbb{Z}$, as follows
\begin{equation}
  \begin{aligned}
    x_0(n+1) & = f \left(  \alpha x_{N-1}(n-1) + \beta M_0 u(n) \right),\\
    x_i(n+1) & = f \left(  \alpha x_{i-1}(n) + \beta M_i u(n) \right),
  \end{aligned}%
  \label{eq:rcevo}%
\end{equation}
where $f$ is a nonlinear function, $u(n)$ is \textcolor{rev2}{the input signal that is injected} into the system, $\alpha$ and $\beta$ are feedback and input gains, respectively, used to adjust the dynamics of the system, and $M_i$ is the input mask, drawn from a uniform distribution over the interval $[-1, +1]$\textcolor{rev2}{, as in e.g. \cite{rodan2011minimum,paquot2012optoelectronic,duport2012all}}. In our implementation, we use a sine function $f = \sin (x)$ as nonlinearity and a ring topology \cite{appeltant2011information,rodan2011minimum} to simplify the interconnection matrix of the network, so that only the first neighbour nodes are connected. Both choices are dictated by the proposed hardware setup \textcolor{rev2}{of the opto-electronic reservoir.} As will be discussed in Sec. \ref{subsec:exp}, we use a light intensity modulator with a sine transfer function as the nonlinear node, and a delay system with time-multiplexing of the reservoir states. A detailed discussion of these experimental aspects can be found in the Supplementary Material of \cite{paquot2012optoelectronic}.

The reservoir computer produces an output signal $y(n)$, given by a linear combination of the states of its internal variables
\begin{equation}
  y(n) = \sum_{i=0}^{N-1} w_i \textcolor{rev2}{(n)} x_i (n),
  \label{eq:rcout}
\end{equation}
where \textcolor{rev2}{$w_i(n)$ are the readout weights, trained either offline (using standard linear regression methods), in which case they are time independent,} or online, as described in the next section, in order to minimise a task-dependent error function, which will be introduced alongside the benchmark tasks further in this paper. \textcolor{rev2}{As discussed below in detail, in hardware implementations of reservoir computing the readout layer can be more complex than Eq. \eqref{eq:rcout}.}

\subsection{Simple gradient descent algorithm}
\label{subsec:grad}

Contrary to offline, or batch learning, where the entire training dataset is used at once to compute the best readout weights $w_i$, online training approach handles the data sequentially in order to optimise the performance step by step. 
As discussed in Sec. \ref{sec:intro}, this allows the Reservoir Computer to be optimised without accurate knowledge of the underlying hardware, which is exactly what is required for an analogue readout layer. This approach can be realised with various algorithms, and in this work we chose to work with the simple gradient descent algorithm, for ease of implementation on the FPGA board.

The gradient, or steepest, descent method is an algorithm for finding a local minimum of a function using its gradient \cite{arfken1985mathematical}. For the task considered here (see the next section) \textcolor{rev2}{we update the readout weights using the procedure given in \cite{bishop2006pattern}}
\begin{equation}
  w_i (n+1) = w_i (n) + \lambda \left( d(n) - y(n) \right) x_i (n),
  \label{eq:wt}
\end{equation}
where $\lambda$ is the step size, used to control the learning rate, and $d(n)$ is the task-dependent target signal (see sections \ref{subsubsec:wc} and \ref{subsubsec:narma}). 
\textcolor{rev2}{The origin of this procedure is that if the error at time $n$ is given by $(d(n)-y(n))^2$, then the derivative of the error with respect to $w_i$ gives $(d(n)-y(n))x_i(n)$, i.e. the right-hand side of Eq. \eqref{eq:wt}.}
At high values of $\lambda$, the weights get close to the optimal values very quickly (in a few steps), but keep oscillating around these values. At low values, the weights converge slowly to the optimal values. In practice, we start with a high value $\lambda=\lambda_0$, and then gradually decrease it during the training phase until a minimum value $\lambda_{min}$ is reached, according to the equation
\begin{equation}
  \lambda(m+1) = \lambda_{min} + \gamma \left( \lambda(m) - \lambda_{min} \right),
  \label{eq:lambda_evo}
\end{equation}
with $\lambda(0) = \lambda_0$ and $m= \left\lfloor n/k \right \rfloor$, where $\gamma<1$ is the decay rate and $k$ is the update rate for the parameter $\lambda$. Previous work has shown that setting $\lambda_0=0.4$, $\lambda_{min}=0$ and $\gamma = 0.999$ is a reasonable choice for good performance \cite{antonik2016online}. The update rate $k$ defines the convergence speed and the resulting error rate : higher $k$ requires a longer training dataset but offers better results.
More complex optimisation methods could be used here, such as simulated annealing \cite{press1986numerical} or stochastic gradient descent \cite{bottou2004stochastic}. However, the above technique is very simple and provides sufficiently good results for this application.

\subsection{Benchmark tasks}

We tested our system on two benchmark tasks commonly used by the reservoir computing community: wireless channel equalisation and emulation of a 10-th order Nonlinear Auto Regressive Moving Average system (NARMA10). Note that, to the best of our knowledge, the latter has never been tested on an online-trained reservoir computer.

\subsubsection{Wireless channel equalisation}
\label{subsubsec:wc}

The operating principle of this practically relevant task is the following. A sequence of symbols $d(n)$ is transmitted through a wireless channel. The receiver records a sequence $u(n)$, which is a corrupted version of $d(n)$. The main sources of corruption are noise (thermal or electronic), multipath propagation, which leads to intersymbol interference, and nonlinear distortion induced by power amplifiers. The goal is to recover $d(n)$ from $u(n)$ \cite{jaeger2004harnessing}.

The channel input signal $d(n)$ contains 2-bit symbols with values picked randomly from $\{-3, -1, 1, 3\}$. The channel is modelled by a linear system with memory of length 10 \cite{mathews1994adaptive}
\begin{align}
  \begin{split}\label{eq:qn}
    q(n) {}& = 0.08 d(n+2) - 0.12 d(n+1) + d(n) \\
           & + 0.18 d(n-1) - 0.1 d(n-2) + 0.091 d(n-3) \\
           & - 0.05 d(n-4) + 0.04 d(n-5) + 0.03 d(n-6) \\
           & + 0.01 d(n-7),
  \end{split}
\end{align}
followed by an instantaneous memoryless nonlinearity
\begin{equation}
  u(n) = q(n) + 0.036 q^2(n) - 0.011q^3(n),
  \label{eq:chan}
\end{equation}
where $u(n)$ is the channel output signal. The reservoir computer has to restore the clean signal $d(n)$ from the distorted noisy signal $u(n)$. The performance is measured in terms of wrongly reconstructed symbols, called the Symbol Error Rate (SER).

\subsubsection{NARMA10}
\label{subsubsec:narma}

The goal of this rather academic task is to emulate a 10-th order Nonlinear Auto Regressive Moving Average system. The input signal $u(n)$ is drawn randomly from a uniform distribution over the interval $[0,0.5]$. The target output $d(n)$ is defined by the following equation
\begin{align}
  \begin{split}\label{eq:narma}
    d(n+1) & = 0.3 d(n) + 0.05 d(n) \left( \sum_{i=0}^9 d(n-i) \right) \\
           & + 1.5 u(n-9) u(n) + 0.1.
  \end{split}
\end{align}
Since the reservoir doesn't produce $d(n)$ exactly, its performance is measured in terms of an error metric. We use the Normalised Mean Square Error (NMSE), given by
\begin{equation}
  \text{NMSE} = \frac{\left\langle \left( y(n) - d(n) \right)^2 \right\rangle}{\left\langle \left( d(n) - \langle d(n) \rangle \right)^2 \right\rangle}.
  \label{eq:mse}
\end{equation}
A perfect match yields $\text{NMSE}=0$, while a completely off-target output gives a NMSE of 1.

\subsection{Proposed Experimental Setup}
\label{subsec:exp}

\begin{figure*}
  \centering
  \includegraphics[width=0.60\textwidth]{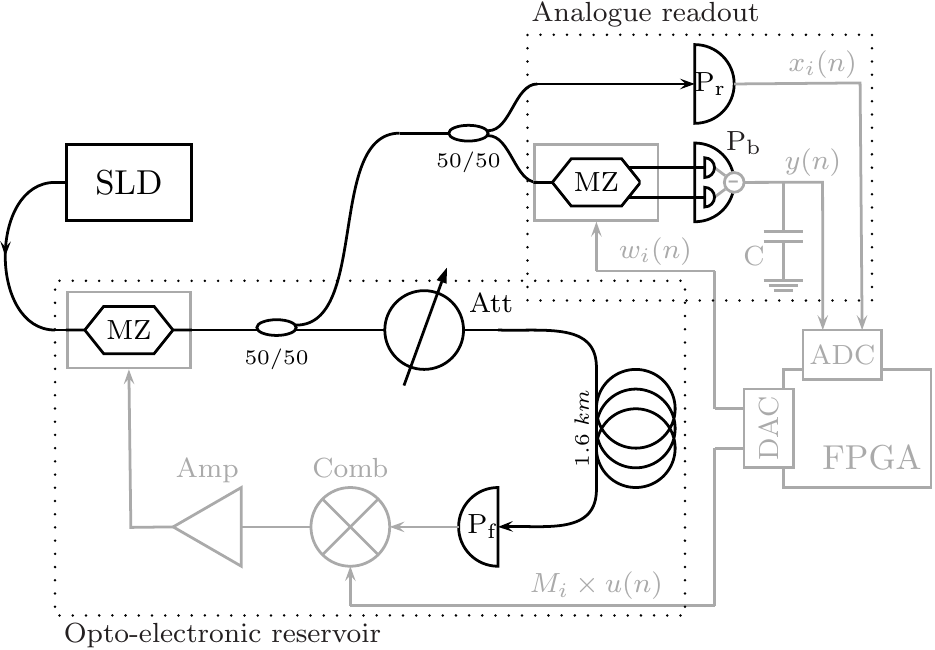}
  \caption{Scheme of the proposed experimental setup. The optical and electronic components are shown in black and grey, respectively. The reservoir layer consists of an incoherent light source (SLD), a Mach-Zehnder intensity modulator (MZ), a $50/50$ beam splitter, an optical attenuator (Att), an approximately $1.6 \units{km}$ fibre spool, a feedback photodiode ($\text{P}_\text{f}$), a resistive combiner (Comb) and an amplifier (Amp). The analogue readout layer contains another $50/50$ beam splitter, a readout photodiode ($\text{P}_\text{r}$), a dual-output intensity modulator (MZ), a balanced photodiode ($\text{P}_\text{b}$) and a capacitor (C). The FPGA board generates the inputs and the readout weights, samples the reservoir states and the output signal, and trains the system.
  }
  \label{fig:exp}
\end{figure*}

Fig. \ref{fig:exp} depicts the proposed experimental setup that we have investigated using numerical simulations.
This section overviews the three main components: the optoelectronic reservoir, the analogue readout layer and the FPGA board. 

\subsubsection{Opto-electronic reservoir}

The optoelectronic reservoir is based on the same scheme as in \cite{paquot2012optoelectronic,larger2012photonic}. 
The reservoir states are encoded into the intensity of incoherent light signal, produced by a superluminiscent diode (Thorlabs SLD1550P-A40). The Mach-Zehnder (MZ) intensity modulator (Photline MXAN-LN-10) implements the nonlinear function, its operating point is adjusted by applying a bias voltage, produced by a Hameg HMP4040 power supply. Half of the signal is extracted from the loop and sent to the readout layer. The optical attenuator (Agilent 81571A) is used to set the feedback gain $\alpha$ of the system (see Eq. \eqref{eq:rcevo}). The fibre spool consists of approximately $1.6 \units{km}$ of single mode fibre, giving a round trip time of $T=7.94 \units{\textmu s}$. The resistive combiner sums the electrical feedback signal, produced by the feedback photodiode (TTI TIA-525I), with the input signal from the FPGA to drive the MZ modulator, with an additional amplification stage of $+27 \units{dB}$ (coaxial amplifier ZHL-32A+) to span the entire $V_\pi$ interval of the MZ modulator.

\subsubsection{Analogue readout layer}
\label{subsubsec:aro}

The analogue readout layer uses the same scheme as proposed in \cite{smerieri2012analog}. The optical power it receives from the reservoir is split in two. Half is sent to the readout photodiode (TTI TIA-525I), and the resulting voltage signal, containing the reservoir states $x_i(n)$, is recorded by the FPGA for the training process (see Eq. \eqref{eq:wt}). The other half is modulated by a dual-output Mach-Zehnder modulator (EOSPACE AX-2X2-0MSS-12) which applies the readout weights $w_i$, generated by the DAC of the FPGA. The outputs of the modulator are connected to a balanced photodiode (TTI TIA-527), which produces a voltage level proportional to difference of the light intensities received at its two inputs. This allows to multiply the reservoir states by both positive and negative weights \cite{smerieri2012analog}. The summation of the weighted states is performed by a low-pass RC filter. The resistance R of the filter, not shown on the scheme, is the $50\;\Omega$ output impedance of the balanced photodiode. The resulting \textcolor{rev2}{output} signal, proportional to $y(n)$, is also recorded by the FPGA, for training and performance evaluation.

Let us compute explicitly the output of the analogue readout layer. The capacitor integrates the output of the balanced photodiode with an exponential kernel and a time constant $\tau$. The impulse response of the RC filter is given in \cite{horowitz1980art}
\begin{equation}
  h(t) = \frac{1}{RC} e^\frac{-t}{RC} = \frac{1}{\tau} e^\frac{-t}{\tau},
\end{equation}
the voltage $Q(t)$ on the capacitor is then given by
\begin{equation}
  Q(t) = \int_{-\infty}^t X(s) W(s) h(t-s) ds,
\end{equation}
where $X(t)$ is the continuous signal, containing the reservoir states, and $W(t)$ are the readout weights, applied to the dual-output intensity modulator. 
\textcolor{rev2}{
The output $y(n)$ is given by the charge on the capacitor at the discrete times $t=nT$:
\begin{equation}
  y(n)=Q(nT).
\end{equation}
Since $X(t)$ and $W(t)$ are stepwise functions $X(t) = x_i(n)$ and $W(t)=w_i$ for $t \in [\theta(i-1), \theta i] $, where $\theta=T/N$ is the duration of one neuron, we can approximate the integration by a discrete summation to obtain
\begin{eqnarray}
  y(n) &=& \theta \sum_{i=1}^N w_i  \left ( \sum_ {k=0}^\infty   x_i(n-k) h(N-i-Nk)\right)\nonumber\\
  &=&  \frac{\theta}{\tau} \sum_{i=1}^N w_i  \left ( \sum_ {k=0}^\infty   x_i(n-k) e^{-\rho(N-i-Nk)}\right)
  \label{eq:qout}
\end{eqnarray}
where we have introduced the RC integrator ratio $\rho = \theta/\tau$.
}

\textcolor{rev2}{The readout layer output $y(t) = Q(t)$ is thus a linear combination of the reservoir states $x_i$, weighted by $w_i$ and by the exponential kernel of the RC filter. 
Note that contrary to usual reservoir computer outputs (see e.g. Eq. (\ref{eq:rcout}), in Eq. (\ref{eq:qout}) the output at time $n$ depends not only on the current states $x_i(n)$, but also on the states at previous times $x_i(n-k)$.}

\textcolor{rev2}{In the previous experimental investigation of the analogue readout \cite{duport2016fully}, the readout weights $w_i$ were computed using ridge regression \cite{tikhonov1995numerical}, assuming an output signal given by Eq. \eqref{eq:rcout}. But since the experiment produced an output similar to Eq. \eqref{eq:qout} instead, the readout weights needed to be corrected appropriately. For more details, we refer to the original paper \cite{duport2016fully}. In the present work, the weights $w_i$ are adjusted gradually to match the reservoir output signal $y(n)$ with the target output $d(n)$ (see Sec. \ref{subsec:grad}), without any assumptions about how these weights actually contribute to the output signal $y(n)$. This is a much easier tasks, which allows to obtain better experimental results, as will be shown in Sec. \ref{sec:res}.
}

\subsubsection{FPGA board}

The reservoir computer is operated by a Xilinx Virtex 6 FPGA chip, placed on a Xilinx ML605 evaluation board. It is paired with a 4DSP FMC151 daughter card, containing one two-channel ADC (Analog-to-Digital converter) and one two-channel DAC (Digital-to-Analog converter). The ADC's maximum sampling frequency is $250 \units{MHz}$ with 14-bit resolution, while the DAC can sample at up to $800 \units{MHz}$ with 16-bit precision. 

The FPGA generates the input signal $M_i \times u(n)$ and sends it into the opto-electronic reservoir. After recording the resulting reservoir states $x_i(n)$ from one delay loop, it executes the simple gradient descent algorithm in order to update the readout weights $w_i(n+1)$. These are sent to the readout layer and used to generate the output signal $y(n)$, also recorded by the FPGA.

\subsection{Numerical simulations}
\label{subsec:numsim}

All numerical experiments were performed in Matlab. We used a custom model of our reservoir computer, based on previous investigations \cite{paquot2012optoelectronic,antonik2016online}, \textcolor{rev2}{that has been shown to emulate very well the dynamics of the real system}. The simulations were performed in discrete time, and took into account the internal structure of the Reservoir Computer described above, such as the ring-like topology, sine nonlinearity and the analogue readout layer with an RC filter. The simulations allow to try out different configurations and to scan various experimental parameters, \textcolor{rev2}{including values} that are impossible to achieve experimentally or imposed by the hardware.
All simulations were performed on a dedicated high-performance workstation with 12-core CPU and $64$ Gb RAM. Since the convergence of the gradient descent algorithm is quite slow, we limited our investigations to a fast update rate $k=10$ (see Eq. \eqref{eq:lambda_evo}), so that each simulation lasted about 24 hours.

The principal goal of the simulations was to check how th online learning approach would cope with experimental difficulties encountered in previous works \cite{smerieri2012analog,duport2016fully}. To that end, we gathered a list of possible issues and scanned the corresponding experimental parameters in order to check the system performance. In particular, we investigated the following parameters:
\begin{itemize}
  \item \textcolor{rev2}{The RC integrator ratio $\rho$.
    This} is the most important parameter of the analogue readout layer. While its accurate measure is not required in our setup -- since we do not correct the readout weights $w_i$ -- it defines the integration span of the filter, and thus the reservoir states that contribute to the output signal. It can thus significantly influence the results. Another question of importance is how dependent the system performance is on the exact value of $\rho$.
  \item The MZ modulator bias. Mach-Zehnder modulators need to be applied a constant voltage to maintain their transfer function symmetric around zero. The devices we were using up to now are subject to slight drifts over time, often resulting in a non-perfectly symmetric response. We thus checked in simulations whether such an offset would impact the results.
  \item The DAC resolution. The precision of the DACs on the FMC151 daughtercard is limited to 16 bits. \textcolor{rev2}{Numerical investigations have shown that the precision of readout weights has a significant impact on the performance, see e.g. \cite{soriano2013optoelectronic,soriano2015delay,antonik2016towards}.}
  We thus checked whether the resolution available is enough for this experiment.
\end{itemize}
Besides these potentially problematic parameters, we also scanned the input and feedback gain parameters (denoted \textcolor{rev2}{by} $\beta$ and $\alpha$ in Eq. \eqref{eq:rcevo}) in order to find the optimal dynamics of the reservoir for each task.

  In a separate set of simulations, we investigated the applicability of the proposed method to nonlinear readout layers. That is, we checked whether the simple gradient descent method would still work with a nonlinear response of the analogue readout layer with respect to the reservoir states $x_i(n)$ (see Eq. \eqref{eq:qout}).
We picked two ``saturation'' functions of sigmoid shape. This choice arises from the transfer function of common light detectors that are linear at lower intensities and become nonlinear at higher intensities. 
We used the following functions: a logistic function, given by 
\begin{equation}
  g_\text{lg}(x) = \frac{2}{1+e^{-2x}} - 1,
  \label{eq:logist}
\end{equation}
and the hyperbolic tangent function, given by
\begin{equation}
  g_\text{ht}(x) = 0.6 \tanh \left( 1.8x \right).
  \label{eq:erf}
\end{equation}
These functions, $g_\text{lg}$ and $g_\text{ht}$, do not model any particular photodiode, but are two simple examples that allow us to address the above question. Both functions are plotted in figure \ref{subfig:nlro}, together with a linear response, for comparison.

We investigated two possible nonlinearities in the output layer. In the first case, the readout photodiode ($\text{P}_\text{r}$ in figure \ref{fig:exp}) produces a nonlinear response, while the balanced photodiode ($\text{P}_\text{b}$ in figure \ref{fig:exp}) remains linear. 
\textcolor{rev2}{This scenario, that we shall refer to as ``nonlinear readout'', allows one to investigate what happens when the reservoir states $x_i$ used to compute the output signal $y(n)$ (see Eq. \eqref{eq:rcout}) differ from those employed to update the readout weights (see Eq. \eqref{eq:wt}). Thus in this case the update rule (Eq. \eqref{eq:wt}) for the output weights  becomes 
\begin{equation}
  w_i (n+1) = w_i (n) + \lambda \left( d(n) - y(n) \right) g(x_i (n)),
  \label{eq:wt-2}
\end{equation}
where $g$ is given by either Eq. \eqref{eq:logist} or Eq. \eqref{eq:erf}, while the output layer is given by Eq. \eqref{eq:qout}.}

\textcolor{rev2}{
In the second case, called ``nonlinear output'', the readout photodiode is linear, but the balanced photodiode exhibits a saturable behaviour. In this case the update rule Eq. \eqref{eq:wt} for the output weights is unchanged, but the output layer Eq. \eqref{eq:qout} becomes 
\begin{equation}
  y(n) =  \frac{\theta}{\tau} \sum_{i=1}^N w_i  \left ( \sum_ {k=0}^\infty   g(x_i(n-k)) e^{-\rho(N-i-Nk)}\right) .
  \label{eq:qout-2}
\end{equation}
}
Note that we have only considered cases with just one nonlinear photodiode, so as to check whether the difference between the reservoir states used for training and those to compute the readout (see Eqs. \eqref{eq:wt} and \eqref{eq:rcout}, respectively) would degrade the performance of the system. The scenario with both nonlinear photodiodes is hence more simple, as the reservoir states are the same in both equations. One could consider the case with two photodiodes exhibiting different nonlinear behaviours. In that situation, similar to the results we will show in Sec. \ref{sec:res}, we expect the algorithm to cope with the difference up to a certain point, before running into troubles. For this reason, we leave that scenario for future investigations.

\section{Results}
\label{sec:res}

\subsection{\textcolor{rev2}{Linear readout: RC circuit}}

In this section we present our numerical results and answer the \textcolor{rev2}{questions} raised in the previous section.

For each of the two tasks considered here, we performed three \textcolor{rev2}{kinds of} simulations: we scanned the RC integrator ratio $\rho = \theta / \tau$ in the first simulation, the MZ bias in the second, and the resolution of the DAC in the third. Furthermore, since different values of these parameters may work better with different dynamics of the reservoir, we also scanned the input gain $\beta$ and the feedback gain $\alpha$ in all three simulations independently, and applied the optimal values in each case.

For both tasks, we used a network with $N=50$ neurons, as in most previous experimental works \cite{paquot2012optoelectronic,duport2012all,vinckier2015high,duport2016fully}. The reservoir was trained on 83000 inputs, with an update rate $k=10$, and then tested over $10^5$ symbols for the channel equalisation task and $10^4$ inputs for NARMA10 task. For statistical purposes, we ran each \textcolor{rev2}{simulation} 10 times, with different random input masks. In the following figures, averaged results over the masks are plotted, while the error bars \textcolor{rev2}{give the standard deviation over the different input masks}. Results related to the channel equalisation task are plotted with solid lines, while dashed lines correspond to those for NARMA10.

\textcolor{rev2}{For the channel equalisation task, our system yields SERs between $10^{-4}$ and $10^{-3}$ depending on the input mask, as summarised in Tab. \ref{tab:nlro} (first line).
This is comparable to previous experiments with the same opto-electronic reservoir:
error rates of order of $10^{-4}$ were reported in \cite{paquot2012optoelectronic} using a digital readout and in \cite{duport2016fully} with an analogue readout, using an RLC filter.} The first experimental analogue system, using a simple RC circuit, as we did in this work, performed significantly worse, with SER of order of $10^{-2}$ \cite{smerieri2012analog}. That is, online learning does not outperform other methods, but allows to obtain significantly better results with a simpler setup.
 
\textcolor{rev2}{As} for the NARMA10 task, we obtain a NMSE of $0.20 \pm 0.02$. 
Previous experiments with a digital readout layer produced $0.168 \pm 0.015 $ \cite{paquot2012optoelectronic} and $0.107 \pm 0.012 $ \cite{vinckier2015high}. With an analogue readout layer, the best NMSE reported was $0.230 \pm 0.023$ \cite{duport2016fully}. Our system thus slightly outperforms the analogue approach, and gets close to the digital one, except for the very good result obtained with a different reservoir, based on a passive cavity \cite{vinckier2015high}. Again, our results were obtained with a simple setup and no modelling of the readout, contrary to \cite{duport2016fully}.

Furthermore, the error rates obtained here can be significantly lowered with more training, as has been demonstrated numerically and experimentally in \cite{antonik2016online}. To keep reasonable simulation times (about 24 hours per simulation), we limited the training to 83000 input values, with an update rate $k=10$. Higher update rates can be used experimentally, since running the opto-electronic setup is much faster than simulating it. We thus expect to obtain lower error rates experimentally with longer training sets and update rates up to $k=200$. To illustrate this point with results reported in \cite{antonik2016online}, short training sets with $k=10$ yielded SERs of order of $10^{-4}$ for the channel equalisation task. Increasing $k$ up to $200$ allowed to decrease the error rate down to $5.7\e{-6}$.

\begin{figure*}
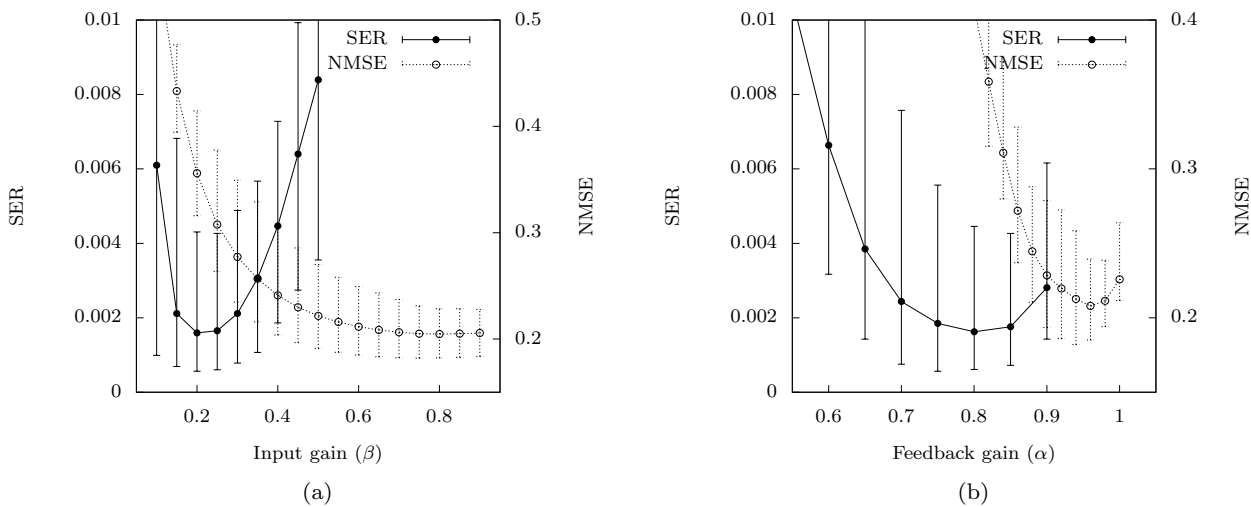

  \centering
  \subfigure[]{\resizebox{0.49\textwidth}{!}{\input{inp.tex}}\label{subfig:inp}}
  \subfigure[]{\resizebox{0.49\textwidth}{!}{\input{fdb.tex}}\label{subfig:fdb}}
  \caption{Reservoir computer performances for different input ($\beta$) and feedback ($\alpha$) gains (solid lines: channel equalisation, dashed lines: NARMA10). 
    \textbf{(a)} While channel equalisation is relatively sensitive to $\beta$, NARMA10 works well in a wide range of values. Note that although it seems that higher input gain would give better results, the dashed curve actually rises slightly \textcolor{rev2}{for large $\beta$, and the optimum input gain is around $0.8$}.
    \textbf{(b)} Both tasks require a system with significant memory (feedback gain at least $\alpha=0.8$), and even a near-chaotic regime for NARMA10 ($\alpha=0.95$).
  }
\end{figure*}

Figures \ref{subfig:inp} and \ref{subfig:fdb} show the influence of input and feedback gain parameters on the performance of the system. All curves present a pronounced minimum, except for the input gain $\beta$ for the NARMA10 task, where values above 0.6 seem to produce comparable results. Note that the channel equalisation task requires a low input signal with $\beta=0.2$, while NARMA10 works best with stronger input and $\beta=0.8$. As for the feedback gain, NARMA10 shifts the system close to the chaotic regime with $\alpha=0.95$, while channel equalisation works better with $\alpha=0.8$.

\begin{figure*}
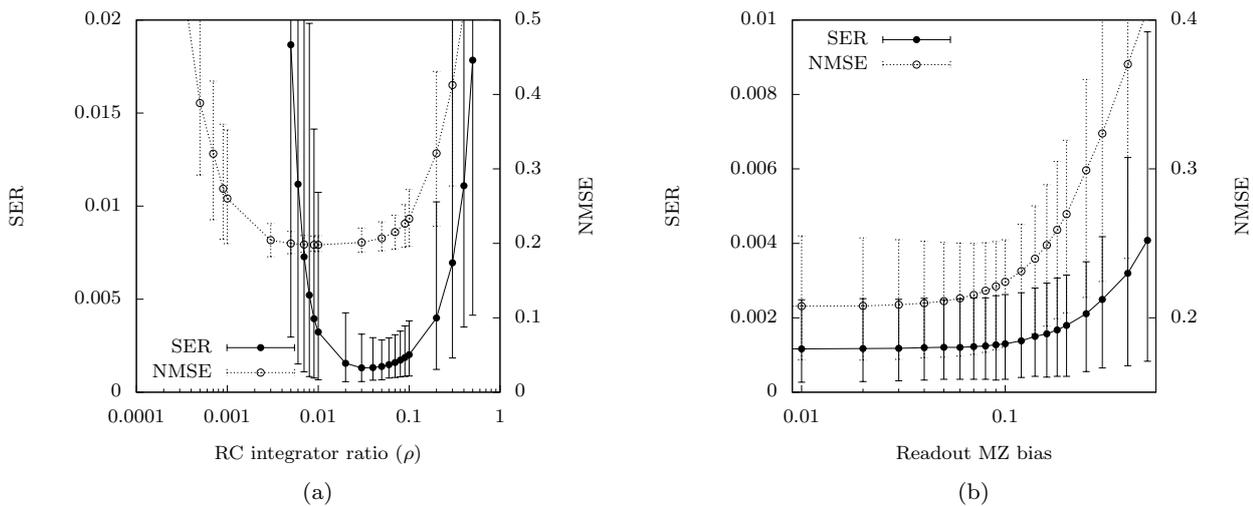

  \centering
  \subfigure[]{\resizebox{0.49\textwidth}{!}{\input{tau.tex}}\label{subfig:tau}}
  \subfigure[]{\resizebox{0.49\textwidth}{!}{\input{mzb.tex}}\label{subfig:mzb}}
  \caption{
    Impact of the RC integrator ratio ($\rho$) and the readout MZ modulator bias on the reservoir computer performance (solid lines: channel equalisation, dashed lines: NARMA10).
    \textbf{(a)} Ratios within $\rho \in [0.03, 0.08]$ are suitable for channel equalisation and $\rho \in [0.002, 0.07]$ for NARMA10. Remarkably, inaccurate choice of $\rho$, and thus $\tau$, will not result in significant performance loss, as long as the value lies approximately in the optimal interval.
    \textbf{(b)} Although the NARMA10 task is more sensitive to this bias, both tasks work reasonably well with a bias up to $0.06$, which is superior to expected experimental deviations.
  }
\end{figure*}

Figure \ref{subfig:tau} shows the results of the scan of the RC integrator ratio $\rho$. Both tasks work well on a relatively wide range of values, with NARMA10 much less sensitive to $\rho$ than channel equalisation. In particular, the channel is equalised best with $\rho=0.03$. With $N=50$, this corresponds to $\tau=T/0.03N = 5.29 \units{\textmu s}$, which is shorter than the roundtrip time $T=7.94\units{\textmu s}$. On the other hand, NARMA10 output is best reproduced with $\rho=0.003$, which yields $\tau=T/0.003N = 52.93 \units{\textmu s}$. This is significantly longer than the roundtrip time $T$, meaning that reservoir states from previous time steps are also taken into account for computation of an output value. This is not surprising, since NARMA10 function has a long memory (see Eq. \eqref{eq:narma}). However, this memory effect in the readout layer is not \textcolor{rev2}{crucial}, as the system performs equally well with higher $\rho$ and thus lower $\tau$. All in all, these results are very encouraging for upcoming experiments, as they show that an accurate choice of capacitor is not crucial for the performance of the system.

Figure \ref{subfig:mzb} illustrates the impact of the bias of the readout Mach-Zehnder modulator on the reservoir computer performance. NARMA10 task is clearly more affected by this offset, as the NMSE grows quickly from a bias of \textcolor{rev2}{roughly} $0.06$. The SER, on the other hand, stays low until $0.1$. For a MZ modulator with $V_\pi = 4.5 \units{V}$ this corresponds to a tolerance of roughly $0.1\units{V}$, which is superior to expected experimental deviations. The Hameg power supply that we use to bias the modulator (see Sec. \ref{subsec:exp}) has a resolution of $0.001\units{V}$.

\begin{figure*}
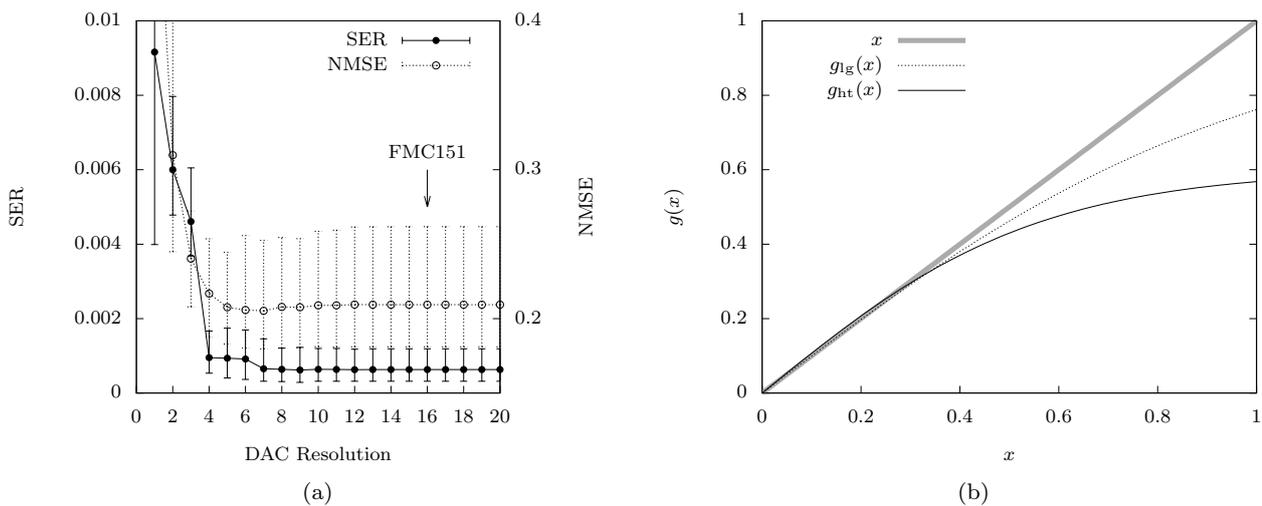

  \centering
  \subfigure[]{\resizebox{0.49\textwidth}{!}{\input{dac.tex}}\label{subfig:dac}}
  \subfigure[]{\resizebox{0.49\textwidth}{!}{\input{nlro.tex}}\label{subfig:nlro}}
  \caption{
    \textbf{(a)}
    Impact of the DAC resolution on the reservoir computer performance (solid lines: channel equalisation, dashed lines: NARMA10). The results show that the 16-bit resolution of the FMC151 daughtercard is sufficient for this application.
    \textbf{(b)}
    Nonlinear response curves of the photodiodes: hyperbolic tangent function $g_\text{ht}$ (solid line) and logistic function $g_\text{lg}$ (dotted line). The linear response is plotted with a thick grey line.
  }
\end{figure*}

Figure \ref{subfig:dac} shows that the 16-bit DAC resolution is not an issue for this experiment, as the minimal precision required for good performance is 8 bits, for both tasks.

\subsection{\textcolor{rev2}{Nonlinear readout}}

\textcolor{rev2}{Table \ref{tab:nlro} sums up} the results obtained with a nonlinear readout layer. We used optimal experimental parameters, as described above, and generated new sets of data for the training and test phases. We investigated two scenarios and used two functions of sigmoid shape, $x \rightarrow g_\text{lg}(x)$ and $x \rightarrow g_\text{ht}(x)$, as described in section \ref{subsec:numsim}. The system was trained over 83000 inputs, with an update rate $k=10$, and tested over $10^5$ symbols for the channel equalisation task and $10^4$ inputs for NARMA10. 
We report error values averaged over 10 trials with different random input masks, as well as the standard deviations. 
The figures show that the performance deterioration is more manifest with the hyperbolic tangent function $g_\text{ht}$, as it is much more nonlinear than the logistic function $g_\text{lg}$. Overall, the added nonlinearity does not have a significant influence on the results in both cases. The SER roughly doubles, at most, for the channel equalisation task. The impact on NARMA10 is barely noticeable, as the error increase of $5\%$ is smaller than the standard deviation. Using offline training on the same system (i.e. with nonlinear output) we observed an increase of the SER by one order of magnitude for the channel equalisation task, and a 30\% increase of the NMSE with the NARMA task. These results show that online training is very well suited for experimental analogue layers, as it can cope with realistic components that do not have a perfectly linear response.

\begin{table}
  \centering
  \begin{tabular}{|l|c|c|c|}
    \hline
    \multirow{2}{*}{Readout} & Transfer & Chan. Equal. & NARMA10 \\
      & function  & (SER $\times 10^{-3}$) & (NMSE) \\
    \hline
    Linear & $x$ & $1.1 \pm 0.7$ & $0.20\pm0.02$ \\[2ex]
    Nonlinear & $g_\text{lg}(x)$ & $ 1.3\pm 0.9$ & $0.21\pm0.03$ \\
    readout & $g_\text{ht}(x)$ & $ 1.2\pm 0.8$ & $0.21\pm0.02$ \\[2ex]
    Nonlinear & $g_\text{lg}(x)$ & $2.0\pm1.6$ & $0.21\pm0.02$ \\
    output & $g_\text{ht}(x)$ & $2.5\pm2.1$ & $0.21\pm0.01$ \\
    \hline
  \end{tabular}
  \caption{Summary of reservoir computer performances with nonlinear readout layers, measured with error metrics related to the tasks considered here. All values are averaged over 10 random input masks and presented with their standard deviations. We used two functions with sigmoid shape to model the response of the photodiodes. We investigated two scenarios: in the ``nonlinear readout'' configuration, the readout photodiode $\text{P}_\text{r}$ is nonlinear, while the balanced photodiode $\text{P}_\text{b}$ is linear, and vice versa in the ``nonlinear output'' scheme. The linear case $x \rightarrow x$ is shown for comparison. For both tasks, the added nonlinearity does not significantly deteriorate the system performance.}
  \label{tab:nlro}
\end{table}

\section{Conclusion}

In this work we propose the online learning technique to improve the performance of analogue readout layers for photonic reservoir computers. We demonstrate an opto-electronic setup with an output layer based on a simple RC filter, and test it, using numerical simulations, on two benchmark tasks. Training the setup online, with a simple gradient descent algorithm, allows to obtain the same level of performance as with a digital readout layer. Furthermore, our approach doesn't require any modelling of the underlying hardware, and is robust against possible experimental imperfections, such as inaccurate choice of parameters or components. It is also capable of dealing with a nonlinearity in the readout layer, such as saturable response of the photodiodes. Finally, we expect the conclusions of the present investigation, namely the advantage of online training, to be applicable to all hardware reservoir computers, and not restricted to the delay dynamical opto-electronic systems used for the sake of illustration in the present work.

Note that the proposed setup is rather slow for practical applications. With a roundtrip time of $T=7.94 \units{\textmu s}$, its bandwidth is limited to $126 \units{kHz}$. This is significantly lower than e.g. a standard Wi-Fi 802.11g channel with a bandwidth of $20\units{MHz}$.
The speed limit of the system is set by the bandwidth of the different components, and in particular of the ADC and DAC. However, the speed of the setup can be easily increased. For instance, with $T=50\units{ns}$ (and thus, a bandwidth of $20\units{MHz}$), and keeping $N=50$, the reservoir states should have a duration of $1\units{ns}$, and hence the ADC and DAC should have bandwidths significantly above $1\units{GHz}$. Such components are readily available commercially. As an illustration of how a fast system would operate, we refer to the optical experiment \cite{brunner2012parallel} in which information was injected into a reservoir at rates beyond $1\units{GHz}$.

The results reported in this work will serve as a basis for future investigations involving experimental validation of the proposed method.
Experimental realisation of an efficient analogue readout layer would allow building fully-analogue high-performance RCs, abandon the slow digital post-processing and take full advantage of the fast optical components. Such setups could be applied to emerging communication channels \cite{bauduin2016high}. Furthermore, fully-analogue setups would open the possibility of feeding the output signal back into the reservoir, thus significantly enriching its dynamics and making it capable of solving signal generation tasks. Recent investigations reported a reservoir computer with digital output feedback capable of periodic and chaotic signal generation \cite{antonik2016towards,antonik2016pattern}. Replacing the digital layer in these implementations with an analogue solution would significantly increase the speed of such generators. Our work thus brings an efficient solution to an important problem in the reservoir computing field, potentially leading to a significant speed gain and a broader range of applications.


\section*{Compliance with Ethical Standards}

\textbf{Funding.} This study was funded by the Interuniversity Attraction Poles program of the Belgian Science Policy Office (grant IAP P7-35 ``photonics@be''), by the Fonds de la Recherche Scientifique FRS-FNRS and by the Action de Recherche Concert\'{e}e of the Acad\'{e}mie Universitaire Wallonie-Bruxelles (grant AUWB-2012-12/17-ULB9).

\noindent
\textbf{Conflict of Interest.} All authors declare that they have no conflict of interest.

\noindent
\textbf{Ethical approval.} This article does not contain any studies with human participants or animals performed by any of the authors.


\end{document}